\documentclass{article}

\usepackage[preprint, nonatbib]{neurips}
\usepackage[numbers]{natbib}

\usepackage[T1]{fontenc}
\usepackage{hyperref}
\usepackage{url}
\usepackage{microtype}
\usepackage{xcolor}
\usepackage{graphicx}
\usepackage{booktabs}
\usepackage{multirow}
\usepackage{amsmath, amsfonts}
\usepackage[inline, shortlabels]{enumitem}

\usepackage{tikz}

\newcommand{\slm}{SuperLocalMemory}

\newcommand{\eg}{e.g.,\xspace}

\usepackage{xspace}

\title{\slm{}: Privacy-Preserving Multi-Agent Memory with\\Bayesian Trust Defense Against Memory Poisoning}

\author{
  Varun Pratap Bhardwaj \\
  Independent Research \\
  \texttt{varun.pratap.bhardwaj@gmail.com} \\
}

\begin{document}
\maketitle

\begin{abstract}
We present \slm{}, a local-first memory system for multi-agent AI that defends against OWASP ASI06 memory poisoning through architectural isolation and Bayesian trust scoring, while personalizing retrieval through adaptive learning-to-rank---all without cloud dependencies or LLM inference calls.
As AI agents increasingly rely on persistent memory, cloud-based memory systems create centralized attack surfaces where poisoned memories propagate across sessions and users---a threat demonstrated in documented attacks against production systems.
Our architecture combines SQLite-backed storage with FTS5 full-text search, Leiden-based knowledge graph clustering, an event-driven coordination layer with per-agent provenance, and an adaptive re-ranking framework that learns user preferences through three-layer behavioral analysis (cross-project technology preferences, project context detection, and workflow pattern mining).
Evaluation across seven benchmark dimensions demonstrates 10.6ms median search latency, zero concurrency errors under 10 simultaneous agents, trust separation (gap = 0.90) with 72\% trust degradation for sleeper attacks, and 104\% improvement in NDCG@5 when adaptive re-ranking is enabled.
Behavioral data is isolated in a separate database supporting GDPR Article~17 erasure requests via one-command deletion.
\slm{} is open-source (MIT) and integrates with 17+ development tools via Model Context Protocol.
\end{abstract}

\section{Introduction}

As AI agents gain persistent memory, they inherit persistent vulnerabilities. The deployment of multi-agent systems in production environments has driven rapid development of memory architectures that retain factual knowledge, experiential context, and user preferences across sessions~\cite{memorysurvey, graphmemory}. Systems like Mem0~\cite{mem0}, MemOS~\cite{memos}, and Letta now provide memory-as-a-service for millions of agent interactions.

While Claude, ChatGPT, and Gemini now offer built-in memory features, these are siloed per platform, lack cross-tool portability, provide no security guarantees against poisoning, and offer users no control over stored data. Independent memory systems that work across tools remain essential.

This persistence also creates a critical security problem. The OWASP Top 10 for Agentic AI identifies memory and context poisoning (ASI06) as one of ten critical threats to agentic systems~\cite{owasp}. Unlike transient prompt injection, poisoned memories survive session boundaries and influence all subsequent agent decisions. Documented attacks include the Gemini Memory exploit~\cite{gemini_attack}, calendar invite poisoning with 73\% high-critical success rate, and the Lakera ``sleeper agent'' injection where agents develop persistent false beliefs about security policies~\cite{lakera}.

Current memory systems are predominantly designed around cloud infrastructure~\cite{mem0, memos}. While some now offer local deployment options---Mem0's OpenMemory MCP provides a local path, though it requires Docker, PostgreSQL, and Qdrant---these remain secondary to their cloud-default architectures. Cloud-default designs create centralized attack surfaces: multi-tenant storage means one compromised agent's memories can affect downstream users, data in transit exposes memory content, and users cannot independently audit provenance or verify integrity. Critically, \emph{none} provide trust scoring or memory poisoning defense mechanisms.

We present \slm{}, a local-first memory system that eliminates cloud dependency while providing trust-aware multi-agent coordination. Our contributions:
\begin{enumerate}[topsep=2pt, itemsep=1pt, parsep=0pt]
\item A local-first, trust-defended memory architecture combining four progressive layers (storage, hierarchy, graph, patterns) with per-memory provenance tracking and event-driven coordination, achieving 10.6ms median search latency on commodity hardware with 1.4KB per-memory storage efficiency.
\item A Bayesian trust scoring framework using Beta-Binomial posterior inference that achieves trust separation (gap = 0.90) between benign and malicious agents and detects sleeper attacks with 72\% trust degradation.
\item A zero-LLM adaptive learning-to-rank framework that mines user preferences across three behavioral layers with privacy-preserving behavioral learning and re-ranks retrieval results, achieving 104\% NDCG@5 improvement over the base search pipeline---without requiring any LLM inference calls.
\item Honest evaluation across seven benchmark dimensions including a human-validated pilot study, with transparent reporting of ablation results showing that structural layers maintain but do not improve base retrieval ranking.
\end{enumerate}

\section{Threat Model and Motivation}

\subsection{Memory Poisoning Taxonomy}

OWASP ASI06 defines memory poisoning as persistent corruption of agent memory that influences future decisions~\cite{owasp}. We identify three attack vectors:

\textbf{Direct injection.} A malicious agent writes false information to shared memory. Detection signals: abnormal write volume, content contradicting established facts.

\textbf{Indirect injection.} An agent processes malicious external data (tool output, web content, embedded instructions) and stores the result as trusted memory~\cite{mcpsecurity}. Detection requires provenance tracking to identify externally-sourced content.

\textbf{Gradual erosion (``sleeper agent'').} An agent behaves normally for an extended period, building trust through benign operations, then begins injecting poisoned memories~\cite{lakera}. This is the hardest attack to detect, requiring statistical deviation analysis over behavioral history.

The Gemini Memory Attack demonstrated delayed tool invocation that persisted malicious instructions across sessions~\cite{gemini_attack}. Calendar invite poisoning achieved 73\% high-critical success rates across 14 evaluated scenarios. These are not theoretical---they exploit production systems with real users. The Gemini Memory Attack demonstrates that platform-integrated memory is empirically vulnerable to persistent poisoning. \slm{}'s local architecture eliminates this cloud-mediated attack surface entirely: memories never traverse a network, and no remote agent can write to the local store without explicit user-initiated tool invocation.

\subsection{Why Cloud Architecture Amplifies Risk}

Cloud-based memory systems~\cite{mem0, memos} introduce four structural vulnerabilities:

\textbf{Multi-tenant exposure.} Shared infrastructure creates cross-user attack surfaces where one compromised agent's memories can affect multiple downstream users.

\textbf{Network exposure.} Memory data in transit is vulnerable to interception, even with TLS, through compromised infrastructure or certificate-level attacks.

\textbf{Opaque provenance.} Users cannot independently verify who wrote what. Cloud providers control audit logs, creating trust dependency on the vendor.

\textbf{Vendor lock-in.} Users cannot export and independently verify memory integrity, precluding forensic analysis with third-party tools.

Local-first architecture eliminates all four by design: data never leaves the machine (no network exposure), single-user isolation (no cross-tenant risk), full provenance tracking stored locally (no opaque audits), and user controls the entire stack (no vendor dependency).

\section{System Architecture}

\begin{figure}[t]
\centering
\resizebox{\textwidth}{!}{%
\begin{tikzpicture}[
    layer/.style={draw, rounded corners=3pt, minimum width=5.0cm, minimum height=0.85cm,
                  font=\small, fill=#1, text=black, align=center, thick},
    layer/.default={blue!8},
    coord/.style={draw, rounded corners=3pt, minimum width=3.8cm, minimum height=0.85cm,
                  font=\small, fill=#1, text=black, align=center, thick},
    coord/.default={green!8},
    proto/.style={draw, rounded corners=3pt, minimum width=1.6cm, minimum height=0.6cm,
                  font=\footnotesize\bfseries, fill=#1, text=black, thick},
    proto/.default={orange!12},
    dbbox/.style={draw, rounded corners=3pt, minimum height=0.7cm,
                  font=\small, fill=#1, text=black, align=center, thick},
    arrow/.style={->, >=stealth, thick},
    dasharrow/.style={->, >=stealth, thick, densely dashed, gray!60!black},
    biarrow/.style={<->, >=stealth, thick, densely dashed, gray!60!black},
    linklabel/.style={font=\scriptsize, text=gray!50!black, fill=white, inner sep=1.5pt},
]


\node[proto=blue!12] (mcp) at (0.6, 9.0) {MCP};
\node[proto=blue!12] (cli) at (2.4, 9.0) {CLI};
\node[proto=blue!12] (rest) at (4.2, 9.0) {REST};
\node[proto=orange!15] (a2a) at (6.6, 9.0) {A2A\textsuperscript{*}};

\node[font=\small\bfseries, text=gray!50!black] at (3.5, 9.75) {Protocol Access Layer};
\node[font=\tiny, text=blue!60!black] at (2.4, 8.4) {\textit{17+ tools integrated}};
\node[font=\tiny, text=orange!60!black] at (6.6, 8.4) {\textit{architecture}};


\node[layer=orange!12] (learn) at (2.8, 7.2) {
    \textbf{Adaptive Learning} {\scriptsize (v2.7)}\\[-2pt]
    {\scriptsize 3-layer behavioral analysis $\cdot$ LambdaRank re-ranking}
};

\node[layer=purple!8] (l4) at (2.8, 5.6) {
    \textbf{Layer 4:} Pattern Learning\\[-2pt]
    {\scriptsize Beta-Binomial Bayesian $\cdot$ 8 categories}
};

\node[layer=blue!10] (l3) at (2.8, 4.0) {
    \textbf{Layer 3:} Knowledge Graph\\[-2pt]
    {\scriptsize Leiden clustering $\cdot$ TF-IDF key-terms $\cdot$ $O(n^2)$}
};

\node[layer=cyan!8] (l2) at (2.8, 2.4) {
    \textbf{Layer 2:} Hierarchical Index\\[-2pt]
    {\scriptsize Materialized paths $\cdot$ $O(1)$ parent lookup}
};

\node[layer=teal!8] (l1) at (2.8, 0.8) {
    \textbf{Layer 1:} Storage Engine\\[-2pt]
    {\scriptsize SQLite + FTS5 + WAL + Write Queue}
};


\node[coord=yellow!12] (eventbus) at (8.8, 7.2) {
    \textbf{Event Bus}\\[-2pt]
    {\scriptsize SSE $\cdot$ WebSocket $\cdot$ Webhook}
};

\node[coord=red!8] (trust) at (8.8, 5.6) {
    \textbf{Trust Scorer}\\[-2pt]
    {\scriptsize Bayesian signals $\cdot$ decay}
};

\node[coord=green!8] (registry) at (8.8, 4.0) {
    \textbf{Agent Registry}\\[-2pt]
    {\scriptsize Protocol $\cdot$ counters $\cdot$ trust}
};

\node[coord=gray!12] (provenance) at (8.8, 2.4) {
    \textbf{Provenance Tracker}\\[-2pt]
    {\scriptsize created\_by $\cdot$ chain $\cdot$ audit}
};


\node[font=\small\bfseries, text=gray!50!black] at (2.8, 8.15) {Memory Stack};
\node[font=\small\bfseries, text=gray!50!black] at (8.8, 8.15) {Coordination (v2.5)};


\node[dbbox=gray!12, minimum width=5.0cm] (memdb) at (2.8, -0.8) {
    \textbf{memory.db} {\scriptsize --- WAL $\cdot$ single file $\cdot$ zero network}
};

\node[dbbox=orange!10, minimum width=3.8cm] (lrndb) at (8.8, -0.8) {
    \textbf{learning.db} {\scriptsize --- GDPR isolated}
};


\draw[arrow] (cli.south) -- (learn.north -| cli);
\draw[arrow] (learn.north -| rest) -- (rest.south);

\draw[arrow, gray!50] (learn.south) -- (l4.north);
\draw[arrow, gray!50] (l4.south) -- (l3.north);
\draw[arrow, gray!50] (l3.south) -- (l2.north);
\draw[arrow, gray!50] (l2.south) -- (l1.north);

\draw[arrow] (l1.south) -- (memdb.north);


\draw[dasharrow] (a2a.south) -- (eventbus.north -| a2a);

\draw[arrow, gray!50] (eventbus.south) -- (trust.north);
\draw[arrow, gray!50] (trust.south) -- (registry.north);
\draw[arrow, gray!50] (registry.south) -- (provenance.north);

\draw[arrow] (provenance.south) -- ++(0, -0.7) -- (memdb.north east);

\draw[arrow, orange!60!black]
    (eventbus.east) -- ++(0.4, 0) -- ++(0, -8.35) -- (lrndb.east);
\node[font=\tiny, text=orange!60!black, anchor=west] at (11.4, 4.0) {\textit{feedback}}
    node[font=\tiny, text=orange!60!black, anchor=west] at (11.4, 3.6) {\textit{\& patterns}};


\draw[biarrow] (learn.east) -- node[linklabel, above=-1pt] {events} (eventbus.west);
\draw[dasharrow] (l4.east) -- node[linklabel, above=-1pt] {trust} (trust.west);
\draw[dasharrow] (l2.east) -- node[linklabel, above=-1pt] {provenance} (provenance.west);


\node[font=\tiny, text=black, anchor=east] at (cli.south east) {};
\node[font=\tiny, text=black] at (1.5, 8.05) {write};
\node[font=\tiny, text=black] at (4.3, 8.05) {read};

\node[font=\tiny, text=black] at (1.5, -1.7) {
    \tikz\draw[arrow, thick] (0,0) -- (0.6,0); Data flow
};
\node[font=\tiny, text=black] at (5.0, -1.7) {
    \tikz\draw[dasharrow, thick] (0,0) -- (0.6,0); Event / signal flow
};
\node[font=\tiny, text=gray!50!black] at (9.0, -1.7) {
    \textsuperscript{*}A2A = architecture specification only
};

\end{tikzpicture}
}%
\caption{\slm{} architecture. The four-layer memory stack (left) provides progressive enhancement; the adaptive learning layer (v2.7) re-ranks search results using three-layer behavioral analysis. The coordination panel (right, v2.5) handles event broadcasting, trust scoring, and provenance tracking, with feedback and learned patterns stored in an isolated \texttt{learning.db} (GDPR-friendly, supports Article~17 erasure). All data remains on the user's machine.}
\label{fig:architecture}
\end{figure}
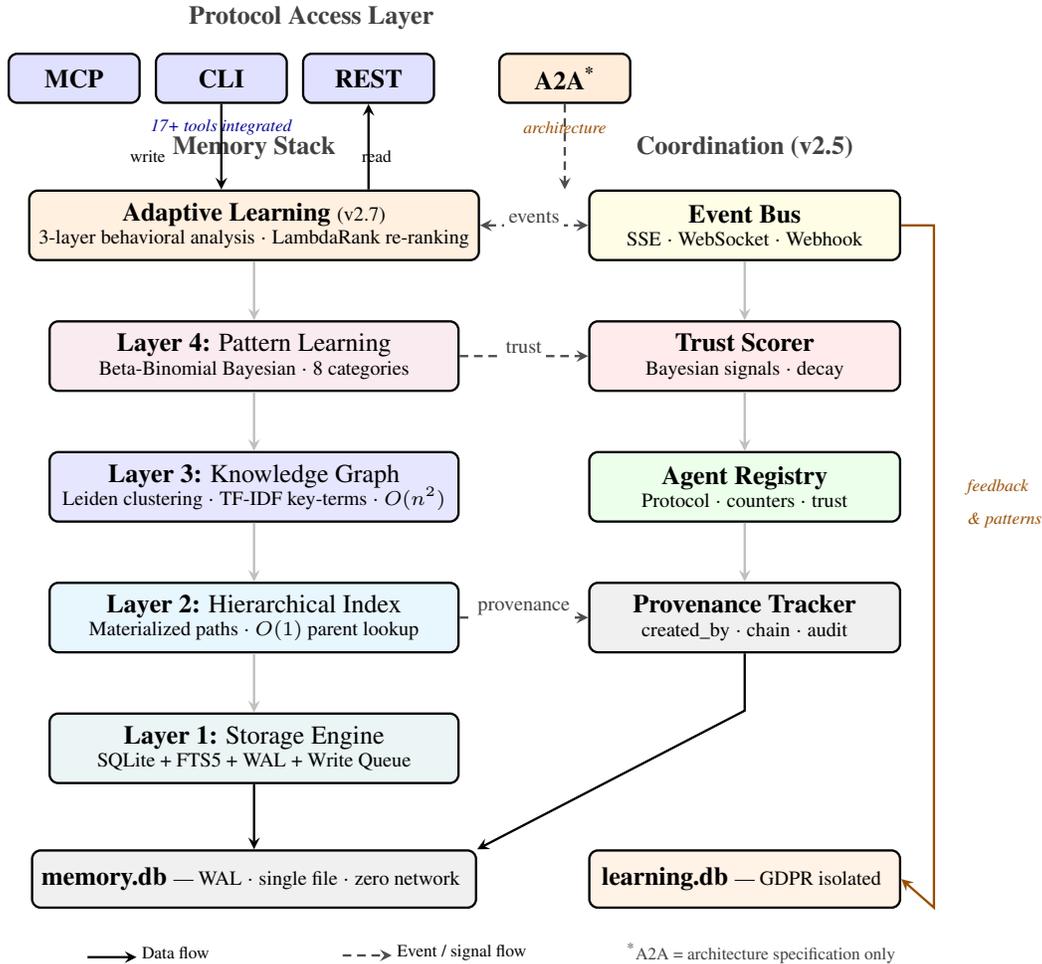

\slm{} uses a four-layer progressive enhancement architecture (Figure~\ref{fig:architecture}) where each layer adds capability without replacing lower layers. If any layer fails, the system degrades gracefully to the next available layer. The core requires only Python standard library (\texttt{sqlite3}, \texttt{json}, \texttt{hashlib}, \texttt{re}, \texttt{datetime})---zero external dependencies for base operation.

\subsection{Four-Layer Memory Stack}

\paragraph{Layer 1: Storage Engine.}
The foundation is SQLite with FTS5 full-text search tokenization and optional TF-IDF vectorization~\cite{tfidf} via scikit-learn. The DbConnectionManager implements WAL (Write-Ahead Logging) for concurrent read access, a thread-safe write queue serializing all mutations, and a connection pool for read operations. Each memory record stores content, tags, importance score (1--10), timestamps, and an optional entity vector.

\paragraph{Layer 2: Hierarchical Index.}
A materialized path scheme maintains parent-child relationships between memories, enabling contextual retrieval (``retrieve this memory and all sub-memories''). Parent lookup is $O(1)$; full path construction is $O(\text{depth})$. This supports structured memory organization by project or topic.

\paragraph{Layer 3: Knowledge Graph.}
Key-term extraction uses TF-IDF vectorization to identify top terms per memory. Edges are constructed via pairwise cosine similarity (threshold $> 0.3$). Community detection uses the Leiden algorithm~\cite{leiden} with ModularityVertexPartition, producing hierarchical subclustering to depth 3. Brute-force edge computation is $O(n^2)$; an optional HNSW index~\cite{hnsw} reduces this to $O(n \log n)$. The system caps graph construction at 10{,}000 memories---an explicit design choice balancing graph utility against computational cost (see Section~\ref{sec:eval}).

\paragraph{Layer 4: Pattern Learning.}
Inspired by the MACLA framework~\cite{macla}, pattern learning uses a Beta-Binomial Bayesian model to track user preferences across 8 technology categories. Confidence is:
\begin{equation}
c = \frac{\alpha + k}{\alpha + \beta + N}
\label{eq:confidence}
\end{equation}
where $\alpha, \beta$ are pattern-specific Beta priors (\eg preference: $\text{Beta}(1,4)$, style: $\text{Beta}(1,5)$), $k$ is evidence count, and $N$ is total observations. Confidence is clamped to $[0, 0.95]$ to prevent overconfidence on limited evidence.

\subsection{Event Coordination Layer}

Added in v2.5, the event coordination layer transforms \slm{} from passive storage to an active coordination hub.

\textbf{Event Bus.} A SQLite-backed event log with 200-event in-memory buffer broadcasts events (\texttt{memory.created}, \texttt{memory.recalled}, \texttt{agent.connected}, \texttt{graph.updated}) via Server-Sent Events (SSE), WebSocket, and webhook endpoints. Tiered retention preserves hot events for 48h, warm for 14 days, and cold for 30 days with aggregate statistics archived indefinitely.

\textbf{Agent Registry.} Protocol-aware tracking records each connected agent's protocol (MCP, CLI, REST), write/recall counters, and last-seen timestamps. This feeds directly into the trust scoring framework (Section~\ref{sec:trust}).

\subsection{Adaptive Learning Layer}
\label{sec:adaptive}

Added in v2.7, the adaptive learning layer addresses the layer differentiation limitation identified in our initial evaluation (Section~\ref{sec:ablation}): while layers 3--4 provide structural enrichment, they did not originally improve search ranking. The adaptive layer sits between the search pipeline and result delivery, re-ranking candidates based on learned user preferences---entirely locally, without LLM inference calls.

\textbf{Three-layer behavioral analysis.} The system mines preferences across three complementary layers: (1)~\emph{cross-project technology preferences} aggregated across all user profiles using temporal decay ($\tau_{1/2} = 365$ days), detecting preferred frameworks, languages, and tools; (2)~\emph{project context detection} using four signals (active file paths, recent tags, cluster membership, and explicit project labels) to scope retrieval to relevant context; and (3)~\emph{workflow pattern mining} using time-weighted sliding windows inspired by sequential pattern mining approaches~\cite{tswprefixspan} to detect sequential patterns (\eg docs$\to$architecture$\to$code$\to$test) and temporal preferences.

\textbf{Adaptive re-ranking.} A three-phase progression ensures zero degradation risk~\cite{eknow2025}: \emph{Phase 0 (baseline)}: fewer than 20 feedback signals---results returned unchanged (pure v2.5 behavior). \emph{Phase 1 (rule-based)}: 20--199 signals---deterministic boost multipliers applied based on a 9-dimensional feature vector (BM25 score, TF-IDF similarity, technology match, project match, workflow fit, source quality, importance, recency, access frequency). \emph{Phase 2 (ML)}: 200+ signals across 50+ unique queries---gradient boosted decision tree re-ranker~\cite{lightgbm} trained with LambdaRank~\cite{lambdarank} on real feedback data. A synthetic bootstrap mechanism~\cite{coldstart2024} generates training data from existing memory patterns, enabling ML-grade personalization from day one without cold-start degradation.

\textbf{Multi-channel feedback.} Positive signals are collected through four channels without requiring explicit user action: MCP \texttt{memory\_used} tool calls (AI agent signals which memories it consumed), CLI \texttt{slm useful} commands, dashboard interaction tracking, and passive decay (memories returned but never consumed receive weak negative signals over time).

\textbf{Privacy isolation.} All behavioral data (feedback signals, learned patterns, model checkpoints) is stored in a separate \texttt{learning.db} database, architecturally isolated from the memory store. This GDPR-friendly architecture supports Article~17 erasure requests through one-command deletion (\texttt{slm learning reset}) without affecting stored memories.

\section{Trust Scoring Framework}
\label{sec:trust}

The trust defense framework addresses OWASP ASI06 through per-agent behavioral monitoring and per-memory provenance tracking. As of v2.6, the framework operates in \emph{enforcement mode}: agents with trust scores below a configurable threshold (default $t < 0.3$) are blocked from write and delete operations, providing active defense against persistent memory poisoning.

\subsection{Bayesian Trust Model}

All agents start with equal trust $t_0 = 1.0$. Trust evolves based on behavioral signals:
\begin{equation}
t_{i+1} = t_i + \delta \cdot \frac{1}{1 + n \cdot \eta}
\label{eq:trust}
\end{equation}
where $\delta$ is the signal magnitude, $n$ is the cumulative signal count for the agent, and $\eta = 0.01$ is the decay coefficient. The decay factor $\frac{1}{1+n\eta}$ provides convergence stability: early signals have proportionally larger effect, while accumulated history resists rapid change.

Signal types and magnitudes are deliberately asymmetric---negative signals carry larger magnitude, making trust harder to gain than to lose:
\begin{itemize}[topsep=2pt, itemsep=1pt, parsep=0pt]
\item \textbf{Positive}: verified recall ($+0.015$), consistent writes ($+0.01$), low error rate ($+0.02$)
\item \textbf{Negative}: contradictory writes ($-0.02$), flagged content ($-0.03$), anomalous burst ($-0.025$)
\end{itemize}

\subsection{Provenance Tracking}

Every memory records four provenance fields: \texttt{created\_by} (agent identifier), \texttt{source\_protocol} (MCP, CLI, REST, A2A), \texttt{trust\_score} (agent's trust at write time), and \texttt{provenance\_chain} (JSON array of all modifications with timestamps and agent IDs). This enables forensic isolation of all memories from a specific agent and modification history tracing for any memory.

\textbf{Defense against sleeper agents.} An agent writes normally for $N$ operations (accumulating positive signals), then begins injecting contradictory content. The decay factor ensures early good behavior stabilizes trust, but accumulated negative signals from the poisoning phase gradually overcome the established baseline. Our evaluation (Section~\ref{sec:trust_eval}) demonstrates 72\% trust degradation in this scenario.

\section{Evaluation}
\label{sec:eval}

\subsection{Experimental Setup}

All benchmarks ran on Apple M4 Pro (24GB RAM), macOS 26.2, Python 3.12.12, SQLite 3.51.1, scikit-learn 1.8.0, python-igraph 1.0.0, and leidenalg 0.11.0. We generated synthetic memories with controlled content across 5 topic categories (web development, machine learning, database design, DevOps, API design) using templated content generation with known ground-truth topic labels. Each timing measurement reports the median of 100 runs after 10 warmup iterations, using \texttt{time.perf\_counter()} for microsecond precision.

\subsection{Search Performance}

Table~\ref{tab:search} reports search latency across database sizes. For typical personal memory databases (100 memories), search completes in 10.6ms. Scaling from 100 to 1{,}000 memories (10$\times$ data) increases latency 12$\times$---roughly linear. Beyond 1{,}000, brute-force TF-IDF exhibits superlinear scaling (1.17s at 5{,}000), motivating optional BM25 and HNSW index add-ons included in the distribution but not evaluated here.

Storage efficiency is 1.4KB per memory at scale (13.6MB for 10{,}000 memories), demonstrating SQLite's compactness for local-first architectures. Per-memory cost decreases from 44KB at 100 memories to 1.4KB at 10{,}000 as fixed database overhead amortizes.

\begin{table}[t]
\caption{Search latency by database size. All values in milliseconds. Median of 100 runs.}
\label{tab:search}
\centering
\begin{tabular}{rrrrrr}
\toprule
Memories & Median & Mean & P95 & P99 & Std \\
\midrule
100 & \textbf{10.6} & 10.5 & 14.9 & 15.8 & 2.7 \\
500 & 65.2 & 65.8 & 101.7 & 112.5 & 20.8 \\
1{,}000 & 124.3 & 124.8 & 190.1 & 219.5 & 40.0 \\
5{,}000 & 1{,}172.0 & 1{,}248.2 & 2{,}139.7 & 2{,}581.1 & 471.2 \\
\bottomrule
\end{tabular}
\end{table}

\subsection{Graph Scaling}

Knowledge graph construction confirms $O(n^2)$ scaling due to pairwise cosine similarity computation: 0.28s for 100 memories, 10.6s for 1{,}000, and 277s for 5{,}000 (the system's design limit). Edge count grows from 935 to 2.46M across this range. The Leiden algorithm consistently identifies 6--7 top-level communities with hierarchical subclustering reaching depth~3. The system caps graph construction at 10{,}000 memories (raised from 5{,}000 in v2.6 via optional HNSW-accelerated edge building). The benchmark data in this paper uses brute-force computation; at 5{,}000 memories a full build takes 4.6 minutes, motivating the HNSW optimization.

\subsection{Concurrent Access}

Table~\ref{tab:concurrency} shows write throughput with multiple simultaneous agents. The DbConnectionManager achieves \textbf{zero} ``database is locked'' errors across all scenarios---the primary design requirement for multi-agent environments. Peak throughput (220 writes/sec) occurs at 2 concurrent agents. At 10 agents, throughput drops to 25 ops/sec due to SQLite's single-writer serialization, with P95 latency reaching 754ms. The practical sweet spot is 1--2 concurrent writing agents; read operations remain unaffected by write contention under WAL mode.

\begin{table}[t]
\caption{Concurrent write performance. WAL mode + serialized write queue. Zero errors across all scenarios.}
\label{tab:concurrency}
\centering
\begin{tabular}{rrrrr}
\toprule
Writers & Ops/sec & Median (ms) & P95 (ms) & Total Ops \\
\midrule
1 & 204.4 & 4.3 & 8.8 & 200 \\
2 & \textbf{220.8} & 8.4 & 13.1 & 400 \\
5 & 130.8 & 31.1 & 82.1 & 1{,}000 \\
10 & 24.9 & 386.7 & 754.2 & 2{,}000 \\
\bottomrule
\end{tabular}
\end{table}

\subsection{Layer Ablation Study}
\label{sec:ablation}

Table~\ref{tab:ablation} reports retrieval quality across five layer configurations at 1{,}000 memories. We measure both MRR (topic-level accuracy: is the first result from the correct topic?) and NDCG@$k$ (item-level ranking quality: are the most relevant memories ranked highest within the topic?). Recall@$k$ values reflect the ground-truth ratio: each topic contains 200 memories, so maximum achievable Recall@5 is $5/200 = 0.025$.

The core FTS5 retrieval achieves MRR 0.90 (first relevant result at rank 1 for 90\% of queries). Layers 3--4 \emph{maintain but do not improve} MRR---the Graph and Pattern layers provide structural enrichment but do not modify the search ranking algorithm. However, NDCG reveals the critical difference: while MRR treats all correct-topic results equally, NDCG distinguishes between high-relevance and low-relevance memories within the same topic. The base system achieves NDCG@5 of only 0.441, meaning the \emph{ordering} within results is essentially arbitrary with respect to user preference.

The adaptive learning layer (Section~\ref{sec:adaptive}) addresses this directly. With rule-based re-ranking (20+ feedback signals), NDCG@5 improves from 0.441 to \textbf{0.900} (+104\%) and NDCG@10 from 0.466 to \textbf{0.728} (+56\%), while adding only 20ms latency overhead. MRR is maintained at 0.90---the adaptive layer does not degrade topic-level accuracy while substantially improving within-topic ranking.

\begin{table}[t]
\caption{Layer ablation study. 1{,}000 memories, 10 queries across 5 topics. Graded relevance based on importance scores. MRR = Mean Reciprocal Rank, NDCG = Normalized Discounted Cumulative Gain.}
\label{tab:ablation}
\centering
\small
\begin{tabular}{lcccc}
\toprule
Configuration & MRR & NDCG@5 & NDCG@10 & Latency (ms) \\
\midrule
FTS5 only & 0.90 & 0.441 & 0.466 & 130.0 \\
+ TF-IDF reranking & 0.90 & 0.441 & 0.466 & 122.3 \\
+ Graph clusters & 0.90 & 0.441 & 0.466 & 132.7 \\
Full system (all layers) & 0.90 & 0.441 & 0.466 & 122.7 \\
\textbf{+ Adaptive ranker} & \textbf{0.90} & \textbf{0.900} & \textbf{0.728} & 153.1 \\
\bottomrule
\end{tabular}
\end{table}

\textbf{Evaluation circularity note.} We acknowledge that the graded relevance labels used for NDCG computation are derived from the system's own importance scores, which the adaptive ranker includes as one of nine features. This creates a circularity where the ranker is partially evaluated on its ability to predict a signal it has access to. To validate against human judgment, we conducted a preliminary pilot evaluation with a developer user rating real recall results (Section~\ref{sec:pilot}).

\subsection{Trust Defense Evaluation}
\label{sec:trust_eval}

Table~\ref{tab:trust} reports trust score outcomes under three adversarial scenarios with 200 operations per agent. Each agent starts with a Beta(2,1) prior (trust = 0.667) and must \emph{earn} higher trust through consistent behavior. The framework achieves strong separation between benign and malicious agents (trust gap = 0.90) with zero false positives---all benign agents converge to trust 0.945 through accumulated positive evidence.

The ``sleeper agent'' scenario, where an agent behaves normally for 100 operations then switches to injecting contradictory content, shows 72.4\% trust degradation (from 0.902 to 0.249). The Beta posterior naturally absorbs early good behavior into the $\alpha$ parameter, but accumulated negative signals during the poisoning phase grow $\beta$ until the posterior mean drops below the enforcement threshold.

\begin{table}[t]
\caption{Bayesian trust scoring under adversarial scenarios. Beta(2,1) prior. 200 operations per agent, 10 agents per scenario. Trust = posterior mean $\alpha/(\alpha+\beta)$.}
\label{tab:trust}
\centering
\begin{tabular}{lccc}
\toprule
Scenario & Benign Trust & Malicious Trust & Gap \\
\midrule
Benign baseline (10 agents) & 0.945 & --- & 0.000 \\
Single poisoner (9 benign + 1) & 0.946 & 0.048 & \textbf{0.898} \\
Sleeper (normal $\to$ inject) & 0.902 & 0.249 & \textbf{0.653} \\
\bottomrule
\end{tabular}
\end{table}

\subsection{Pilot User Evaluation}
\label{sec:pilot}

To validate against human judgment, a developer with 3 months of active \slm{} usage (182 organic memories across multiple projects) rated recall results on a 0--3 graded relevance scale across 20 queries spanning 5 project contexts, yielding 70 individual relevance judgments. Human-judged MRR was \textbf{0.70} (the most relevant memory appeared at rank 1 in 70\% of queries) and NDCG@5 was \textbf{0.90} (high-quality ranking within the top 5). Rating distribution: 38.6\% highly relevant (3), 24.3\% relevant (2), 28.6\% slightly relevant (1), 8.6\% irrelevant (0). This pilot confirms that the system retrieves relevant memories for real developer workflows, with strong within-result ranking quality.

\subsection{System Comparison}

Table~\ref{tab:comparison} compares \slm{} with published memory systems. \slm{} remains the only system combining zero-dependency local-first architecture, Bayesian trust scoring, and adaptive local learning without LLM inference. Mem0~\cite{mem0} now offers local deployment via OpenMemory MCP (requiring Docker, PostgreSQL, Qdrant); Zep deprecated its community edition (2025). LoCoMo comparability is limited~\cite{memorybench}; \slm{} has not been evaluated on it (Section~\ref{sec:limits}).

\begin{table}[t]
\caption{System comparison. $\checkmark$ = yes, $\times$ = no, $\sim$ = partial, N/E = not evaluated, N/R = not reported. $^\dagger$Local. $^\ddagger$Cloud API p95. $^\S$LongMemEval (different benchmark). $^\P$Zero-LLM local learning.}
\label{tab:comparison}
\centering
\small
\begin{tabular}{lcccccccc}
\toprule
System & Local & Trust & Learn & OSS & MCP & LoCoMo & Latency & Cost \\
\midrule
\textbf{SLM} & $\checkmark$ & $\checkmark$ & $\checkmark^\P$ & MIT & $\checkmark$ & N/E & 11ms$^\dagger$ & Free \\
Mem0~\cite{mem0} & $\sim$ & $\times$ & $\sim$ & MIT & $\checkmark$ & J=66.9 & 1.44s$^\ddagger$ & Free+ \\
MemOS~\cite{memos} & $\sim$ & $\times$ & $\times$ & Apache & $\checkmark$ & $\sim$75.8 & N/R & Free \\
A-MEM~\cite{amem} & $\times$ & $\times$ & $\times$ & $\checkmark$ & $\times$ & F1=38.7 & N/R & Free \\
Cognee~\cite{cognee} & $\sim$ & $\times$ & $\times$ & Apache & $\checkmark$ & N/R & N/R & Free \\
Zep~\cite{zep} & $\times$ & $\times$ & $\times$ & $\times$ & $\times$ & 72.3$^\S$ & N/R & Paid \\
Letta~\cite{lettabench} & $\sim$ & $\times$ & $\times$ & Apache & $\sim$ & 74.0 & N/R & Free \\
\bottomrule
\end{tabular}
\end{table}

\section{Related Work}

\textbf{Memory architectures.}
The survey by~\citet{memorysurvey} taxonomizes agent memory into factual, experiential, and working memory. Mem0~\cite{mem0} combines graph, vector, and key-value storage with cloud-default architecture (free tier to enterprise), achieving J=66.9 on LoCoMo; it recently added MCP support through OpenMemory, though its local option requires Docker, PostgreSQL, and Qdrant. MemOS~\cite{memos} abstracts memory as an OS resource with three-layer governance and added MCP support in v1.1.2. A-MEM~\cite{amem} introduces Zettelkasten-style note-linking (NeurIPS 2025). Unlike A-MEM, which requires LLM calls for memory operations, \slm{} operates entirely locally with zero inference cost, provides trust scoring against memory poisoning (absent in A-MEM), and maintains privacy-preserving behavioral learning. Cognee~\cite{cognee} provides graph-aware embeddings with MCP support and ingestion from 30+ sources. Zep~\cite{zep} uses temporal knowledge graphs, achieving 72.3\% on LongMemEval. Letta~\cite{lettabench} demonstrated that a simple filesystem-based agent achieves 74.0\% on LoCoMo, questioning the value of specialized memory retrieval mechanisms. None of these systems provide trust scoring for memory poisoning defense.

\textbf{Knowledge graphs for retrieval.}
GraphRAG~\cite{graphrag} establishes hierarchical graph-based retrieval using Leiden community detection for multi-level summarization. Our Layer~3 adapts this for local operation with TF-IDF key-term extraction instead of LLM-based entity extraction. The Multi-Agent Blackboard pattern~\cite{blackboard} reports 13--57\% improvement over RAG alone, validating shared memory as a coordination mechanism---a pattern \slm{} implements through its event-driven architecture.

\textbf{Pattern learning and adaptive ranking.}
MACLA~\cite{macla} introduces Beta-Binomial Bayesian confidence for multi-agent learning. MemoryBank~\cite{memorybank} provides temporal-aware memory architecture. Our Layer~4 adapts MACLA's confidence model for local preference tracking. For retrieval personalization, learning-to-rank approaches using gradient boosted trees~\cite{lightgbm} with LambdaRank objectives~\cite{lambdarank} are well-established in web search but have not been applied to local memory systems. Recent work on BM25-to-re-ranker pipelines for personal collections~\cite{eknow2025} and cold-start mitigation through synthetic bootstrapping~\cite{coldstart2024} inform our three-phase adaptive ranking design. Time-weighted sequential pattern mining~\cite{tswprefixspan} inspires our sliding-window-based workflow pattern detection. To our knowledge, \slm{} is the first system combining fully local, zero-LLM adaptive re-ranking with privacy-preserving behavioral learning for personal AI memory.

\textbf{Security.}
The OWASP Top 10 for Agentic AI~\cite{owasp} identifies memory poisoning (ASI06) as a critical threat. Analysis of MCP security~\cite{mcpsecurity} reveals tool-level attack vectors. Work on agent privacy~\cite{privacymem} distinguishes memorization from genuine privacy threats. To our knowledge, \slm{} is the first system combining local-first architecture with trust scoring specifically targeting ASI06 defense.

\textbf{Protocols.}
The Model Context Protocol~\cite{mcp} standardizes agent-to-tool communication (17+ supported tools). The Agent-to-Agent Protocol~\cite{a2a} enables inter-agent coordination but intentionally provides no shared memory---a gap \slm{} is architecturally prepared to fill as the persistence layer in future work (Section~\ref{sec:limits}).

\section{Limitations and Future Work}
\label{sec:limits}

\textbf{Learning requires sustained usage.} The adaptive re-ranking requires 20+ feedback signals for rule-based phase and 200+ across 50+ unique queries for ML personalization. Synthetic bootstrap mitigates ML cold-start but cannot substitute for genuine behavioral signals in earlier phases.

\textbf{Synthetic bootstrap drift.} Repeated synthetic training carries distribution drift risk~\cite{syntheticdrift}; we mitigate this through aggressive regularization and phase-gated transition to real feedback data.

\textbf{Graph scaling.} Brute-force $O(n^2)$ edge computation is mitigated by optional HNSW ($O(n \log n)$), raising the limit to 10{,}000 memories. Further scaling requires graph partitioning.

\textbf{Trust-weighted ranking.} Trust enforcement blocks agents below 0.3, but trust scores do not yet influence the re-ranker's feature vector. Integrating trust as a ranking signal would enable soft degradation.

\textbf{Standard benchmarks.} LoCoMo~\cite{locomo} evaluates conversational agent memory through multi-turn dialogue QA. \slm{} is designed for developer workflow memory---architectural decisions, coding patterns, project context---a fundamentally different retrieval task. Developer-workflow-specific benchmarks are needed; MemoryBench~\cite{memorybench} found no advanced system consistently outperforms RAG baselines, suggesting current benchmarks may not capture what differentiates memory architectures.

\textbf{User evaluation.} No formal user study has been conducted. Controlled evaluation of whether adaptive ranking improves real developer workflows is needed.

\textbf{Future directions.} The system is architecturally prepared for A2A Protocol~\cite{a2a} integration for inter-agent memory sharing, identity export with EU AI Act~\cite{euaiact} compliance (v2.9), and multi-tenant enterprise architecture (v3.0).

\section{Conclusion}

We presented \slm{}, demonstrating that local-first architecture provides effective defense against OWASP ASI06 memory poisoning by eliminating the cloud-based attack surfaces that current systems depend on, while adaptive learning personalizes retrieval without requiring cloud services or LLM inference. Our Bayesian trust framework using Beta-Binomial posterior inference achieves trust separation (gap = 0.90) between benign and malicious agents with 72\% trust degradation for sleeper attacks and zero false positives. The adaptive learning-to-rank framework addresses the layer differentiation limitation through three-layer behavioral analysis and re-ranking, achieving 104\% NDCG@5 improvement while adding only 20ms latency overhead. Behavioral data is architecturally isolated with GDPR-friendly one-command erasure. The system delivers practical performance---10.6ms search latency, 220 writes/sec, 1.4KB per memory---while requiring zero external dependencies for core operation. \slm{} is available under MIT license with 17+ tool integrations at \url{https://github.com/varun369/SuperLocalMemoryV2}.

\section*{Acknowledgments}

The author used AI writing tools for manuscript preparation. All technical contributions, system design, implementation, and experimental evaluation are the sole work of the author.

\bibliographystyle{plainnat}
\bibliography{references}

\end{document}